\documentclass[10pt,twocolumn,letterpaper]{article}

\usepackage{cvpr}
\usepackage{times}
\usepackage{epsfig}
\usepackage{graphicx}
\usepackage{amsmath}
\usepackage{amssymb}
\usepackage{comment}
\usepackage{multirow}
\usepackage{xcolor}

\usepackage[pagebackref=true,breaklinks=true,letterpaper=true,colorlinks,bookmarks=false]{hyperref}

\cvprfinalcopy 

\definecolor{lightgray}{RGB}{220,220,220}
\definecolor{darkblue}{RGB}{0,0,127}
\definecolor{darkgreen}{RGB}{0,127,0}
\definecolor{darkred}{RGB}{200,0,0}

\ifcvprfinal\fi

\usepackage[affil-it]{authblk}

\usepackage{enumitem}

\title{The 4th AI City Challenge}

\begin{document}

\pagenumbering{gobble}

\author{
Milind Naphade$^1$ \hspace{0.9cm}
Shuo Wang$^1$ \hspace{0.9cm}
David C. Anastasiu$^2$ \hspace{0.9cm}
Zheng Tang$^8$\\
Ming-Ching Chang$^3$ \hspace{0.9cm}
Xiaodong Yang$^9$ \hspace{0.9cm}
Liang Zheng$^4$ \hspace{0.9cm}
Anuj Sharma$^5$\\
Rama Chellappa$^6$ \hspace{0.9cm}
Pranamesh Chakraborty$^7$
} 
\affil{ 
$^1$ NVIDIA Corporation, CA, USA  \hspace{0.8cm} 
$^2$ Santa Clara University, CA, USA \\ 
$^3$ University at Albany, SUNY, NY, USA  \hspace{0.8cm} 
$^4$ Australian National University, Australia \\
$^5$ Iowa State University, IA, USA  \hspace{0.8cm} 
$^6$ University at Maryland, College Park, MD, USA \\ 
$^7$ Indian Institute of Technology Kanpur, India  \hspace{0.8cm} 
$^8$ Amazon, WA, USA  \hspace{0.8cm} 
$^9$ QCraft, CA, USA \\
}

\maketitle

\begin{abstract}

The AI City Challenge was created to accelerate intelligent video analysis that helps make cities smarter and safer. Transportation is one of the largest segments that can benefit from actionable insights derived from data captured by sensors, where computer vision and deep learning have shown promise in achieving large-scale practical deployment. The 4th annual edition of the AI City Challenge has attracted 315 participating teams across 37 countries, who leveraged city-scale real traffic data and high-quality synthetic data to compete in four challenge tracks. Track 1 addressed video-based automatic vehicle counting, where the evaluation is conducted on both algorithmic effectiveness and computational efficiency. Track 2 addressed city-scale vehicle re-identification with augmented synthetic data to substantially increase the training set for the task. Track 3 addressed city-scale multi-target multi-camera vehicle tracking. Track 4 addressed traffic anomaly detection. The evaluation system shows two leader boards, in which a general leader board shows all submitted results, and a public leader board shows results limited to our contest participation rules, that teams are not allowed to use external data in their work. The public leader board shows results more close to real-world situations where annotated data are limited. Our results show promise that AI technology can enable smarter and safer transportation systems. 


\end{abstract}

\section{Introduction}

Transportation is one of the largest segments that can benefit from actionable insights derived from data captured by sensors. However, difficulties including poor data quality, the lack of annotations, and the absence of high-quality models are some of the biggest impediments to unlocking the value of the data \cite{BigData:Challenge:IEEEAccess2017}. The AI City Challenge was first launched in 2017 to accelerate the research and development in Intelligent Transportation Systems (ITS) by providing access to massive amounts of labeled data to feed learning-based algorithms. We shared a platform for participating teams to innovate and address real-world traffic problems, as well as evaluated their algorithms against common datasets and metrics. The past three annual editions~\cite{Naphade17AIC17, Naphade18AIC18, Naphade19AIC19} of the challenge have witnessed major impact in research areas of traffic, signaling systems, transportation systems, infrastructure, and transit. 

The 4th edition of the challenge is organized as a workshop at CVPR 2020, which has pushed the development of ITS in two new ways. First, the challenge introduced a track that not only measured effectiveness on tasks that were relevant to transportation but also measured the efficiency of completing these tasks and the ability of systems to operate in real time. To the best of our knowledge, this is the first such challenge that combines effectiveness and efficiency evaluation of tasks needed by the Department of Transportation (DOT) for operational deployments of these systems. The second change was the introduction of augmented synthetic data for the purpose of substantially increasing the training set for the task of re-identification (ReID). The four tracks for the challenge are listed as follows:

\begin{itemize}[leftmargin=12pt] 

\item \textbf{Turn-counts for signal timing planning:} This task counts four-wheel vehicles and freight trucks that follow pre-defined movements from multiple camera scenes. The dataset contains 31 video clips (about 9 hours in total) captured from 20 unique camera views. 

\item \textbf{Vehicle ReID with real and synthetic training data:} This task is tested against the \textit{CityFlow-ReID} benchmark~\cite{Tang19CityFlow}, where teams perform vehicle ReID based on vehicle crops from multiple cameras placed at multiple intersections. A synthetic dataset~\cite{Yao19VehicleX, Tang19PAMTRI} with more than 190,000 images of over 1,300 distinct vehicles forms an augmented training set to be used along with the real-world data.

\item \textbf{City-scale multi-target multi-camera vehicle tracking:} In this task, teams perform multi-target multi-camera (MTMC) vehicle tracking, which is evaluated on the \textit{CityFlow} benchmark~\cite{Tang19CityFlow}. We introduced a new test set for the challenge this year that contains over 200 annotated vehicle identities across nearly 12,000 frames. 

\item \textbf{Traffic anomaly detection:} This task evaluates methods on a dataset provided by the DOT of Iowa. Each participating team submits at most 100 anomalies detected, including wrong turns, wrong driving direction, lane change errors, and all other anomalies, based on video feeds available from multiple cameras at intersections and along highways.  

\end{itemize}

We had over 1,100 total submissions to the evaluation system ($\S$~\ref{sec:eval}) across the four challenge tracks. When submitting results, teams could choose to submit to the \textit{Public} or the \textit{General} leader boards. As the name suggests, the \textit{Public} leader board has been shared with the public, where the submissions compete for the challenge prizes. We enforce two rules for \textit{Public} leader board contest: (1) Teams may not use external data in computing their prediction models for any of the tracks. (2) Teams must submit their code, models, and any labels they created on the training datasets to the competition organizers before the end of the challenge. Alternatively, teams could submit to the the \textit{General} leader board, which ranks all submissions, including the \textit{Public} leader board submissions.

We have seen strong participation in the past three editions of the AI City Challenge. Statistics of the 4th AI City Challnege show growing impact among academic and industrial research communities. This year, we have 315 participating teams composed of 811 individual researchers from 37 countries. We received 233, 258, 239, and 224 requests, respectively, for participating in the challenge tracks. From these, 93 of the teams signed up for an evaluation system account, out of which 76 and 55 individual teams submitted results to the \textit{General} and \textit{Public} leader boards, respectively. 

This paper summarizes the 2020 AI City Challenge preparation and results. In the following sections, we describe the challenge setup ($\S$~\ref{sec:challenge:setup}), challenge data preparation ($\S$~\ref{sec:dataset}), evaluation methodology ($\S$~\ref{sec:eval}), analysis of submitted results ($\S$~\ref{sec:results}), and a brief discussion of insights and future trends ($\S$~\ref{sec:discussion}).

\section{Challenge setup}
\label{sec:challenge:setup}

We have set up the 4th edition of the AI City Challenge with similar rules as the previous ones, where teams are allowed to participate in one or more of the four challenge tracks. In terms of the time-frame, we made the training and testing data available to participants in early January 2020. Due to the new publication rules of CVPR, the 4th AI City Challenge was scheduled to finish on April 9, 2020 (a month earlier than the previous editions). In order to be considered as prize contenders, teams were requested to submit both training and testing code, additional labels, and generated models for validation of their performance on the leader boards. 

For all the data made available to the participating teams, we have taken extra attention in redacting private information such as human faces and license plates. The tasks in the four challenge tracks are elaborated as follows.

\textbf{Track 1: Multi-class multi-movement vehicle counting.} 
Participating teams were asked to count four-wheel vehicles and freight trucks that follow pre-defined movements from multiple camera scenes. Teams performed vehicle counting separately for left-turning, right-turning and through traffic at a given intersection approach. This helps traffic engineers understand the traffic demand and freight ratio on individual corridors. The developed capabilities can be used to design better intersection signal timing plans and improve traffic congestion mitigation. To maximize the practical value of the outcome from this track, both the vehicle counting effectiveness and the module running efficiency were considered as a weighted sum towards the final score for each team. The team with the highest final score will be declared the winner of this track.

\textbf{Track 2: Vehicle ReID with real and synthetic training data.} 
Participating teams were challenged for vehicle ReID based on image crops from different camera perspectives. This task is critical for algorithms to learn fine-grained appearance features that distinguish vehicles, even those of the same color, model, and year. In this year's challenge, the training set was composed of both real-world data and synthetic data. The use of synthetic data was encouraged as they can be simulated under various environments and can produce large-scale training sets by applying domain adaptation. The team with the highest accuracy in identifying vehicles among the top $K$ matches of each query will be selected as the winner.

\textbf{Track 3: City-scale MTMC vehicle tracking.}
The task for participating teams was to track vehicles across multiple cameras both at a single intersection and across multiple intersections in a city. Results can be used by traffic engineers to model journey times along entire corridors. The team with the best accuracy in detecting vehicles and recovering their trajectories across multiple cameras/intersections will be declared as the winner.

\textbf{Track 4: Traffic anomaly detection.}
Based on more than 50 hours of videos collected from different camera views at multiple freeways by the DOT of Iowa, each participating team was asked to submit a list of at most 100 detected anomalies. The anomalies include single and multiple vehicle crashes and stalled vehicles. Regular congestion was not considered as an anomaly. The team with the highest average precision and the most accurate anomaly start time prediction in the submitted events will be the winner of this track.

\section{Datasets}
\label{sec:dataset}

Data for this challenge comes from multiple traffic cameras from a city in the United States as well as from state highways in Iowa. Specifically, we have time-synchronized video feeds from several traffic cameras spanning major travel arteries of the city. Most of these feeds are high resolution 1080p feeds at 10 frames per second. The vantage point of these cameras is for traffic and transportation purposes and the data have been redacted in terms of faces and license plates to address data privacy issues. In addition to the datasets used in the previous AI City Challenges, this year we added a new vehicle counting dataset and a a synthetic vehicle dataset. 

Specifically, the datasets provided for the challenge this year were \textit{CityFlow}~\cite{Tang19CityFlow, Naphade19AIC19} (for Track 2 - ReID and Track 3 - MTMC tracking), \textit{VehicleX}~\cite{Yao19VehicleX, Tang19PAMTRI} (for Track 2 - ReID), Iowa DOT~\cite{Naphade18AIC18} dataset (for Track 4 - anomaly event detection and Track 1 - vehicle counting). 

\subsection{The {\bf \textit{CityFlow}} dataset}

\begin{figure}[t]
\centerline{
  \includegraphics[width=1.0\linewidth]{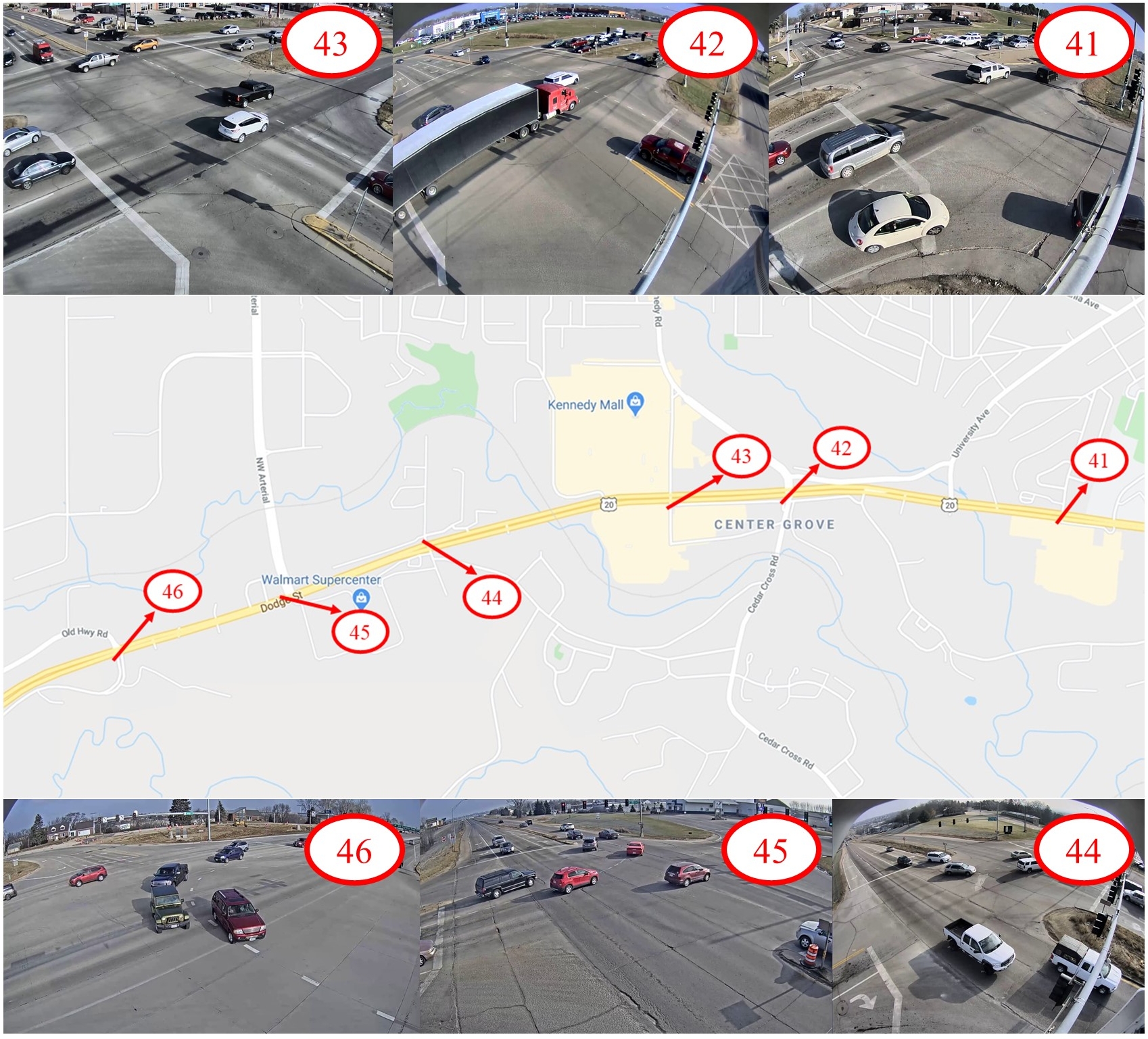}
\vspace{-0.1cm}
}
\caption{The {\bf \textit{CityFlow}} benchmark~\cite{Tang19CityFlow} captured at multiple intersections along a city highway. Here six new test camera views are shown.}
\label{fig:mtmc_test_set}
\vspace{-0.4cm}
\end{figure}

Similar to the AI City Challenge in 2019, the \textit{CityFlow} benchmark~\cite{Tang19CityFlow, Naphade19AIC19} was adopted for the tasks of ReID and MTMC tracking. The dataset consists of nearly 3.5 hours of synchronized videos captured from multiple vantage points at various urban intersections and along highways. Videos are 960p or better, and most have been captured at 10 frames per second. To prevent teams from overfitting the test data provided in the previous edition, we have made the original test set into a validation set, and launched a new test set for the challenge this year. Included in the new test set are six simultaneously recorded videos all captured from different intersections along a city highway with nearly 12,000 frames and over 200 annotated vehicle identities. The geo-locations of the six cameras and example frames are presented in Fig.~\ref{fig:mtmc_test_set}. 

In total, the dataset contains 215.03 minutes of videos collected from 46 cameras spanning 16 intersections in a mid-sized U.S. city. The distance between the two furthest simultaneous cameras is 4 km. The dataset is divided into six scenarios. Of these, three are used for training, two are used for validation, and the remaining one is used for testing. The entire dataset contains nearly 300K bounding boxes for 880 distinct annotated vehicle identities. Only vehicles passing through at least two cameras have been annotated. Additionally, in each scenario, the offset from the start time is available for each video, which can be used for synchronization. We also provided the teams the baseline camera calibration and single-camera tracking results, which can be leveraged for spatio-temporal association of vehicle trajectories. 

A subset of the \textit{CityFlow} dataset, a.k.a. \textit{CityFlow-ReID}, is reserved for the ReID task in Track 2. There are 666 total vehicle identities, where half of them are used for training, and the other half for testing. The training and test sets contain 36,935 and 18,290 vehicle crops, respectively. And we have 1,052 image queries to be identified in the test set. The evaluation and visualization tools are available with the dataset package for teams to measure their performance quantitatively and qualitatively. 

\subsection{The {\bf \textit{VehicleX}} dataset}

\begin{figure}[t]
\centerline{
\includegraphics[width=1.0\linewidth]{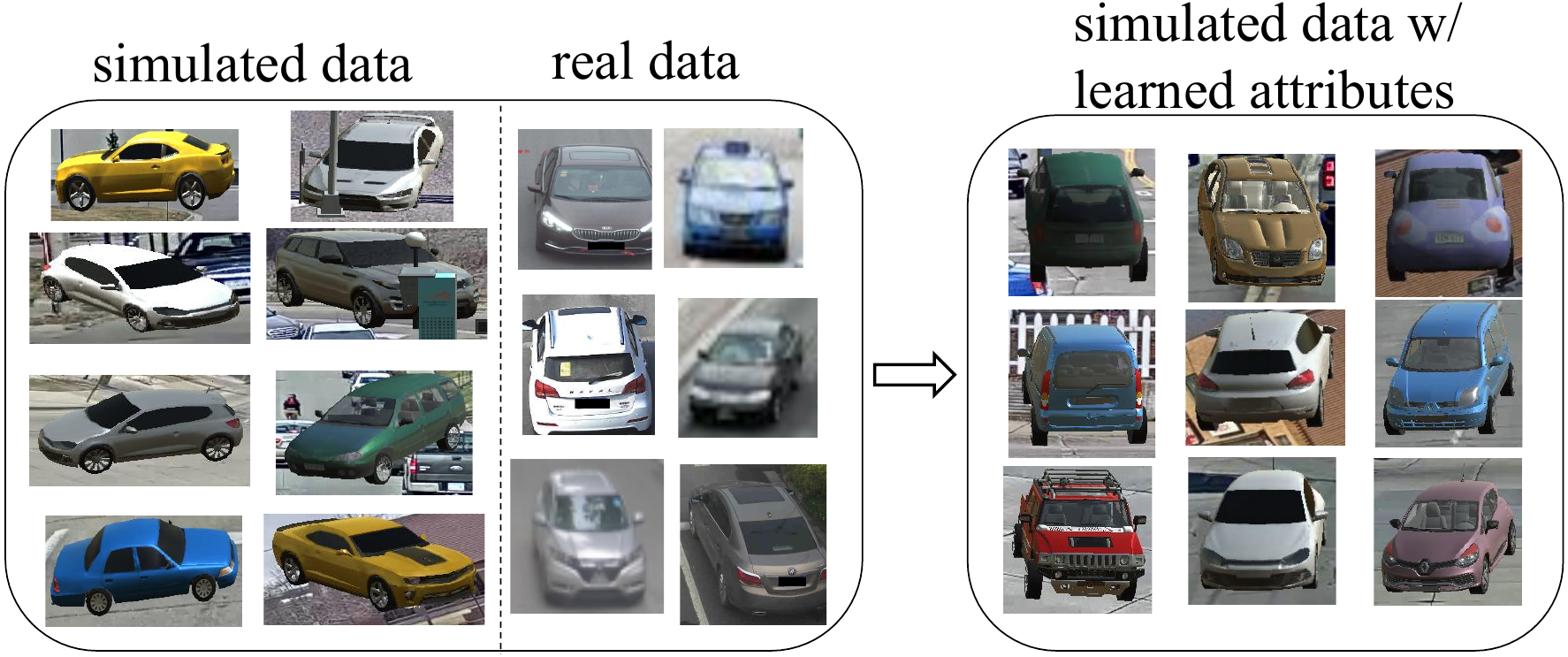}
\vspace{-0.1cm}
}
   \caption{The {\bf \textit{VehicleX}} dataset contains synthetic training data through domain adaptation that can effectively reduce the content gap with the real data for vehicle ReID. }
\label{fig:vehiclex}
\vspace{-0.4cm}
\end{figure}

The \textit{VehicleX} dataset \cite{Yao19VehicleX, Tang19PAMTRI} as shown in Fig.~\ref{fig:vehiclex} is a large-scale public 3D vehicle dataset containing high-quality synthetic images rendered on real-world backgrounds for vehicle ReID use. It can be used for the joint training with detection and tracking datasets (\emph{i.e.,} \textit{Cityflow-ReID}) to improve the real-world ReID performance. \textit{VehicleX} contains more than 190,000 images from 1,362 vehicle identities. Each vehicle identity corresponds to a 3D model with editable attributes including the viewpoint, lighting and rendering conditions. 

In order to minimize the domain gap between synthetic and real-world data, an {\em attribute descent} approach is used to edit the synthetic dataset to make the appearance similar to real-world datasets in terms of key attributes such as the viewpoint \cite{Yao19VehicleX}. The \textit{Unity} engine draws random images from the \textit{Cityflow} dataset to be used as the backgrounds in the attribute descent. Moreover, SPGAN~\cite{deng2018image} is used to adapt the style of synthetic image to match the real-world style. The above methods can significantly reduce the content discrepancy between simulated and real data, thereby making \textit{VehicleX} look visually plausible and similar to the real-world vehicles cropped from natural images. We also provided the \textit{Unity} engine which links the Python API to participating teams, so the teams can potentially create more synthetic data if needed. The detailed annotations including car types and color are provided in the \textit{VehicleX} dataset. With the large number of images, vehicle types, colors, and the comprehensive attribute annotations, this dataset can benefit large-scale ReID for the research community. 


\subsection{Vehicle counting dataset}

\begin{figure}[t]
\centerline{
  \includegraphics[width=1.0\linewidth]{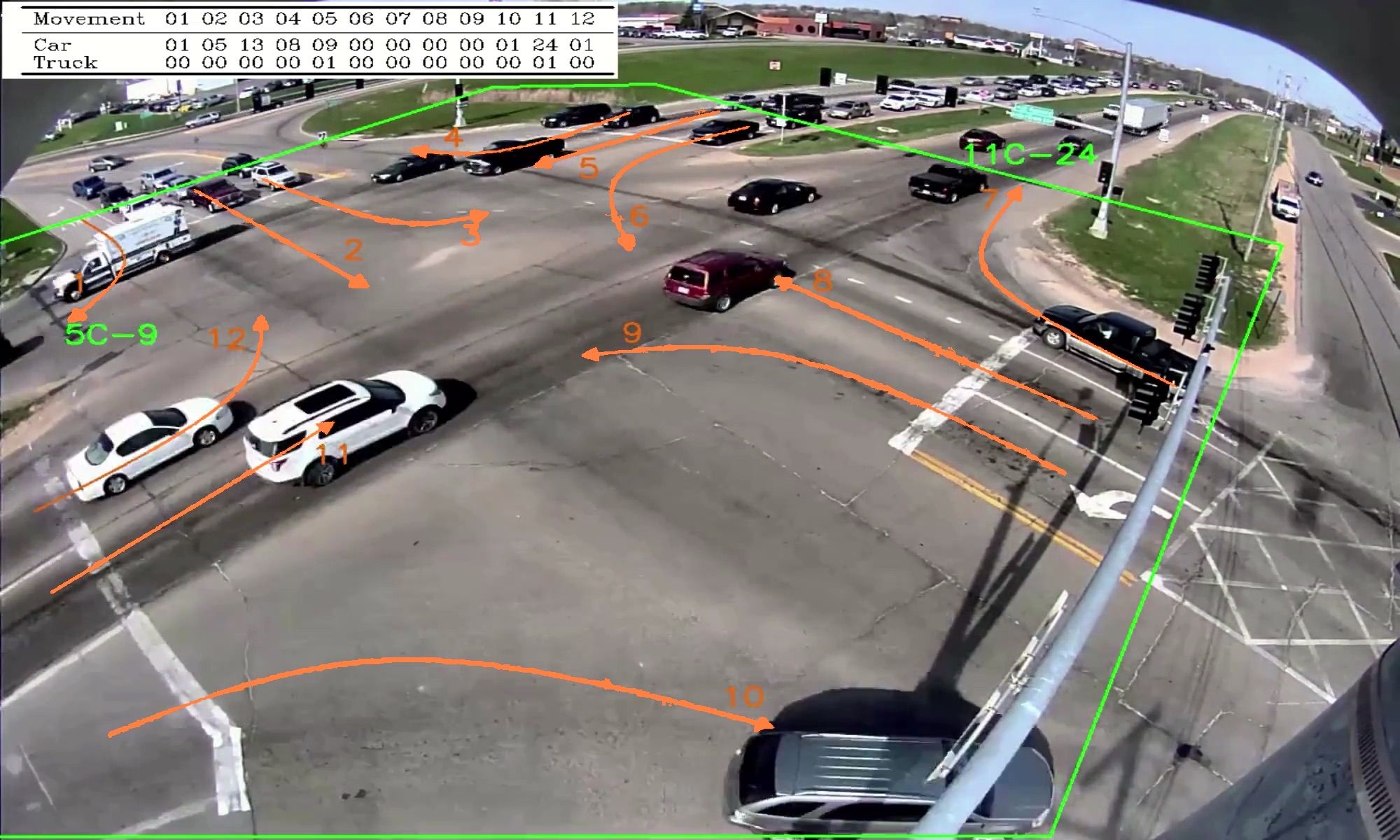}
\vspace{-0.0cm}
}
\caption{The {\bf vehicle counting dataset} designed for multi-class, multi-movement vehicle counting.}
\label{fig:turn_counts}
\vspace{-0.4cm}
\end{figure}

The vehicle counting data set contains 31 video clips (about 9 hours in total) captured from 20 unique camera views. Some cameras provide multiple video clips to cover different lighting and weather conditions. Videos are 960p or better, and most have been captured at 10 frames per second. Detailed documents describing the Region of Interest (ROI) and the Movements of Interest (MOI) that are relevant to the vehicle counting task setup in each camera view are also provided. Fig.~\ref{fig:turn_counts} provides an example view for vehicle counting, where the ROI is marked in a green polygon and the MOIs are marked in the set of orange arrows. 
The ROIs and MOIs are defined to remove the ambiguity that whether a certain vehicle should be counted or not especially near the start and end of a video segment. Any vehicle presented in the ROI becomes a candidate to be counted and a certain candidate should be counted at the moment of fully exiting the ROI if its movement is one of the pre-defined MOIs. By following these predefined ROI and MOI rules, two people manually counting the same video should yield the same result. 
The ground truth counts for all videos were manually created following these rules. In this contest, {\em cars} and {\em trucks} were counted separately for each MOI as shown in Fig.~\ref{fig:turn_counts}. Sedan car, SUV, van, bus, and small trucks such as pickup trucks, and UPS mail trucks were counted as ``cars''. Medium trucks such as moving trucks, garbage trucks, and large trucks such as tractor trailers and 18-wheelers were counted as ``trucks''. The ground truth counts were cross-validated manually by multiple annotators.

\subsection{Iowa DOT anomaly dataset}

\begin{figure}[t]
\centerline{
  \includegraphics[width=1.0\linewidth]{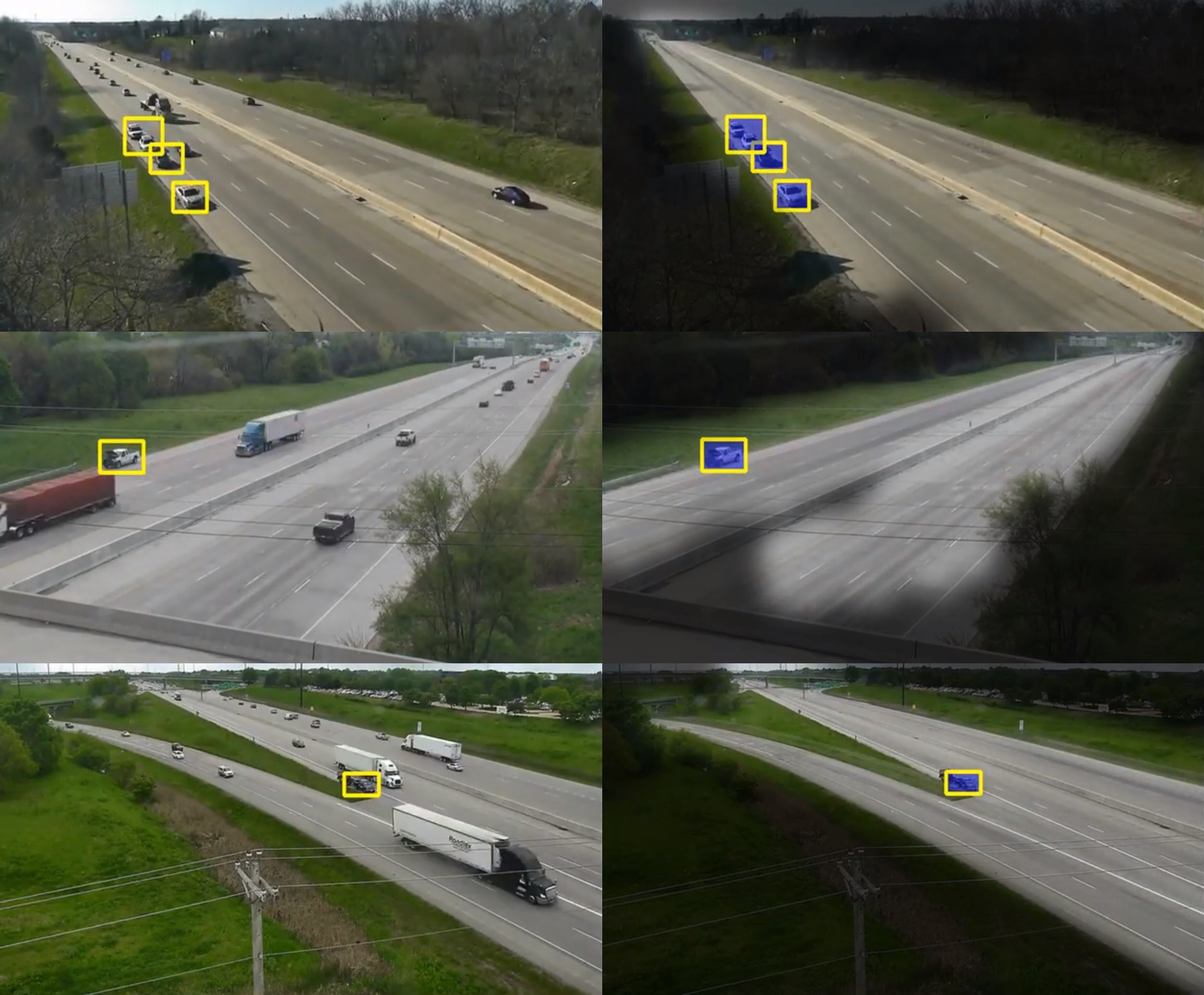}
\vspace{-0.0cm}
}
\caption{The {\bf traffic anomaly dataset} containing traffic anomalies caused by vehicle crashes and stalled vehicles. The left column shows detected anomalies in the original frames. The right column presents background modeling results~\cite{Nguyen19Vehicle}.}
\label{fig:anomaly}
\vspace{-0.4cm}
\end{figure}

The Iowa DOT anomaly dataset consists of 100 video clips each in the training and test datasets. The clips were recorded at 30 frames per second at a resolution of $800 \times 410$. Each video clip is approximately 15 minutes in duration and may include a single or multiple anomalies. However, if a second anomaly is reported while the first anomaly is still in progress, it is counted as a single anomaly. The traffic anomalies consist of single and/or multiple vehicle crashes and stalled vehicles (see Fig.~\ref{fig:anomaly}~\cite{Nguyen19Vehicle}). A total of 21 such anomalies were presented in the training dataset across 100 clips. Unlike previous editions of the AI City Challenge, the participating teams were not allowed to use any external dataset for training and validation except for ImageNet- or COCO-based pre-trained object detection models. 

\section{Evaluation methodology}
\label{sec:eval}

Similar to previous AI City Challenges~\cite{Naphade18AIC18,Naphade19AIC19}, we allowed teams to submit multiple runs for each track to an \textbf{online evaluation system} that automatically measured the effectiveness of results upon submission, which encouraged teams to continue to improve their results until the end of the challenge. Teams were allowed a maximum of 5 submissions per day and a maximum number of submissions for each track (20 for Tracks 2 and 3, and 10 for Tracks 1 and 4). Submissions that lead to a format or evaluation error did not count against a team's daily or maximum submission totals. 

To further encourage competition among the teams, the evaluation system showed not only a team's own performance, but also the top-3 best scores on the leader boards (without revealing identifying information for those teams). To discourage excessive fine-tuning to improve performance, the results shown to the teams prior to the end of the challenge were computed on a 50\% subset of the test set for each track. After the challenge submission deadline, the evaluation system revealed the full leader boards with scores computed on the entire test set for each track.

Teams competing for the challenge prizes were not allowed to use external data or manual labeling to fine-tune their models' performance, and their results were published on the \textit{Public} leader board. For the first time this year, we allowed teams using additional external data or manual labeling to also submit results, which were published on a separate \textit{General} leader board.

\subsection{Track 1 evaluation}

The Track 1 evaluation score (S1) is a weighted combination between the Track 1 efficiency score ($S1_{\text{efficiency}}$) and the Track 1 effectiveness score ($S1_{\text{effectiveness}}$). 

\begin{equation}\label{s1}
\begin{aligned}
S1 = \alpha S1_{\text{efficiency}} + \beta S1_{\text{effectiveness}},\\
\text{where}\, \alpha = 0.7, \beta=0.3.
\end{aligned}
\end{equation}

The $S1_{\text{efficiency}}$ score is based on the total Execution Time provided by the contestant, adjusted by an Efficiency Base factor, and normalized within the range [0, 5x video play-back time].
$S1_{\text{efficiency}} = 1 - \frac{\text{time} \times \text{base factor}}{5 \times \text{video total time}}$. The Efficiency Base factor is computed as the ratio between the execution time of a subset of the {\em pyperformance}~\footnote{\url{https://pyperformance.readthedocs.io/}} benchmark on the user's system and on a baseline system.

The $S1_{\text{effectiveness}}$ score is computed as a weighted average of normalized weighted root mean square error score (nwRMSE) across all videos, movements, and vehicle classes in the test set, with proportional weights based on the number of vehicles of the given class in the movement. To reduce jitters due to labeling discrepancies, each video is split into \textit{k} segments and we consider the cumulative vehicle counts from the start of the video to the end of each segment. The small count errors that may be seen in early buckets due to counting before or after the segment breakpoint will diminish as we approach the final segment. The nwRMSE score is the weighted RMSE (wRMSE) between the predicted and true cumulative vehicle counts, normalized by the true count of vehicles of that type in that movement. If the wRMSE score is greater than the true vehicle count, the nwRMSE score is assigned 0, else it is (1-wRMSE/vehicle count). To further reduce that impact of errors on early segments, the wRMSE score weighs each record incrementally in order to give more weight to recent records. 
\vspace{-0.2cm}
\begin{equation}\label{wRMSE}
\begin{aligned}
wRMSE = \sqrt{\sum_{i=1}^n w_i (\hat{x}_i - x_i)^2}, \\
\text{where} \;w_i = \frac{i}{\sum_{j=1}^n j} = \frac{2i}{n(n+1)}.
\end{aligned}
\end{equation}

Since the contestants could have reported inaccurate efficiency scores, competition prizes will only be awarded based on the scoring obtained when executing the teams' codes on the held out Track 1 Dataset B. To ensure comparison fairness, Dataset B experiments will be executed on the same server. Additionally, teams with anomalous efficiency scores on Dataset A will be disqualified.

\subsection{Track 2 evaluation} 
\label{sec:track2:eval}

In Track 2, given the large size of \textit{CityFlow-ReID}, we used the rank-$K$ mAP metric to measure performance, which computes the mean of the average precision (the area under the Precision-Recall curve) over all the queries when considering only the top-$K$ results for each query ($K=100$). In addition to the rank-$K$ mAP results, our evaluation server also computes the rank-$K$ Cumulative Matching Characteristics (CMC) scores for $K\in\{1,5,10,15,20,30,100\}$, which are popular metrics for person ReID evaluation. While these scores were shared with the teams for their own submissions, they were not used in the overall team ranking and were not displayed in the leader boards.

\subsection{Track 3 evaluation}
\label{sec:track1:eval}

The primary task of Track 3 was identifying and tracking vehicles that traveled through the viewpoints of at least two of the cameras in the \textit{CityFlow} dataset. As in 2019, we adopted the IDF1 score~\cite{Ristani16Performance} from the \textit{MOTChallenge}~\cite{Bernardin2008,Li09Learning} to rank the performance of each team. The score measures the ratio of correctly identified detections over the average number of ground-truth and computed detections. In the multi-camera setting, the score is computed in a video made up of the concatenated videos from all cameras. The ground truth consists of the bounding boxes of multi-camera vehicles labeled with a consistent global ID. A high IDF1 score is obtained when the correct multi-camera vehicles were discovered, accurately tracked within each video, and labeled with a consistent ID across all videos in the dataset. For each submission, the evaluation server also computes several other performance measures, including ID match precision (IDP), ID match recall (IDR), and detection precision and recall. While these scores were shared with the teams for their own submissions, they were not used in the overall team ranking and were not displayed in the leader boards.

\subsection{Track 4 evaluation}
\label{sec:track4:eval}
Track 4 performance is measured by combining the detection performance and detection time error. Specifically, the Track 4 score ($S_4$), for each participating team, is computed as
\begin{equation}\label{s3}
S_4 = F_1 \times (1-NRMSE^t),
\end{equation}
where the $F_1$ score is the harmonic mean of the precision and recall of anomaly prediction. For video clips containing multiple ground-truth anomalies, credit is given for detecting each anomaly. Conversely, multiple false predictions in a single video clip are counted as multiple false alarms. If multiple anomalies are provided within the time span of a single ground-truth anomaly, we only consider the one with minimum detection time error and ignore the rest. We expect all anomalies to be successfully detected and penalize missed detection and spurious ones through the $F_1$ component in the $S_4$ evaluation score. We compute the detection time error as the RMSE between the ground-truth anomaly start time and predicted start time for all true positives. To obtain a normalized evaluation score, we calculate $NRMSE^t$ as the normalized detection time RMSE using min-max normalization between 0 and 300 frames (for videos of 30 frames per second, this corresponds to 10 seconds), which represents a reasonable range of RMSE values for the anomaly detection task. Specifically, $NRMSE^t$ of team $i$ is computed as
\vspace{-0.3cm}
\begin{equation}\label{NRMSE}
NRMSE_i^t = \frac{\min(RMSE_i, 300)}{300}.
\end{equation}

\section{Challenge results}
\label{sec:results}

Tables~\ref{table:1},~\ref{table:2},~\ref{table:3}, and~\ref{table:4} summarize the leader boards for Track 1 (turn-counts for signal timing planning), Track 2 (vehicle ReID), Track 3 (city-scale MTMC vehicle tracking), and Track 4 (traffic anomaly detection) challenges, respectively. \textit{General} indicates general submissions. 

\subsection{Summary for the Track 1 challenge}

\begin{table}[t]
\caption{Summary of the Track 1 leader board.}
\label{table:1}
\centering
\footnotesize
\begin{tabular}{|c|c|c|c|}
\hline
Rank & Team ID & Team name (and paper) & Score \\
\hline\hline
1 & 99 & Baidu~\cite{Baidu20Count} & 0.9389 \\
\hline
2 & 110 & ENGIE~\cite{ENGIE20Count} & 0.9346 \\
\hline
3 & 92 & CMU~\cite{CMU20Count} & 0.9292 \\
\hline
6 & 74 & BUT~\cite{BUT20Count} & 0.8829 \\
\hline
7 & 6 & KISTI~\cite{KISTI20Count} & 0.8540 \\
\hline
9 & 80 & HCMUS~\cite{HCMUS20CountReIDAnomaly} & 0.8064 \\
\hline
13 & 75 & UAlbany~\cite{UAlbany20CountReIDMTMCTAnomaly} & 0.3116 \\
\hline\hline
N/A (\textit{General}) & 60 & DiDi~\cite{DiDi20Count} & 0.9260 \\
\hline
N/A (\textit{General}) & 108 & VT~\cite{VT20Count} & 0.8138 \\
\hline
\end{tabular}
\vspace{-0.4cm}
\end{table}

All submitted teams follow a similar three-step strategy in tackling the vehicle counting task: (1) vehicle detection, (2) vehicle tracking, and (3) movement assignment from trajectory modeling and classification. 

For vehicle detection, 
most teams~\cite{KISTI20Count, VT20Count, BUT20Count} selected YOLOv3~\cite{Redmon18YOLOv3} pre-trained on COCO as the primary detector, while some others~\cite{CMU20Count, UAlbany20CountReIDMTMCTAnomaly} selected Mask R-CNN. 
CenterNet was used in \cite{HCMUS20CountReIDAnomaly}, and a comprehensive comparison study was performed in \cite{DiDi20Count}, in which NAS-FRP combined with the GMM background model was ultimately used.
Faster R-CNN~\cite{Ren15} was used by the top two teams~\cite{Baidu20Count, ENGIE20Count}. 

For vehicle tracking, DeepSORT~\cite{Wojke17} was most widely used by teams~\cite{KISTI20Count, VT20Count, BUT20Count, DiDi20Count, Baidu20Count}. The UAlbany team \cite{UAlbany20CountReIDMTMCTAnomaly} adopted Hungarian matching algorithm to associate detections into tracklets, considering both spatial and appearance features. The team from HCMUS~\cite{HCMUS20CountReIDAnomaly} showed that the IoU-based tracking was simple yet very effective. The CMU team~\cite{CMU20Count} used a newly proposed tracking algorithm that can be processed in real time. The team from ENGIE~\cite{ENGIE20Count} defined a final loss function based on vehicle counting results from motion-based tracking that were optimized for each camera.

For movement assignment, several strategies are developed, which can be organized into two categories:
(1) Manually defined movement ROIs, where some teams defined the ROI using a single zone or a tripwire~\cite{HCMUS20CountReIDAnomaly, ENGIE20Count, UAlbany20CountReIDMTMCTAnomaly}, and other teams~\cite{KISTI20Count, BUT20Count} represented the movements with a pair of enter/exit zones.
(2) Data-driven movement ROI, based on the similarity between query and modeled trajectories. The CMU team~\cite{CMU20Count} manually created the modeled trajectories, while the others~\cite{VT20Count, CMU20Count, DiDi20Count, Baidu20Count} created the modeled trajectories by clustering a set of selected trajectories. 
In all cases, teams developed effective techniques that can merge broken trajectories and reduce identity switches using various filtering and smoothing methods. 

\subsection{Summary for the Track 2 challenge}

\begin{table}[t]
\caption{Summary of the Track 2 leader board.}
\label{table:2}
\centering
\footnotesize
\begin{tabular}{|c|c|c|c|}
\hline
Rank & Team ID & Team name (and paper) & Score \\
\hline\hline
1 & 73 & Baidu-UTS~\cite{Baidu20ReID} & 0.8413 \\
\hline
2 & 42 & RuiyanAI~\cite{RuiyanAI20ReID} & 0.7810 \\
\hline
3 & 39 & ZJU~\cite{ZJU20ReID} & 0.7322 \\
\hline
4 & 36 & Fraunhofer~\cite{Fraunhofer20ReID} & 0.6899 \\
\hline
7 & 72 & UMD~\cite{UMD20ReIDMTMCTAnomaly} & 0.6668 \\
\hline
15 & 38 & NTU~\cite{NTU20ReID} & 0.5781 \\
\hline
19 & 9 & BUPT~\cite{BUPT20ReID} & 0.5354 \\
\hline
20 & 35 & TUE~\cite{TUE20ReID} & 0.5166 \\
\hline
26 & 80 & HCMUS~\cite{HCMUS20CountReIDAnomaly} & 0.3882 \\
\hline
27 & 85 & Modulabs~\cite{Modulabs20ReID} & 0.3835 \\
\hline
30 & 66 & UAM~\cite{UAM20ReID} & 0.3623 \\
\hline\hline
N/A (\textit{General}) & 87 &  CUMT~\cite{CUMT20ReID} & 0.6656 \\
\hline
N/A (\textit{General}) & 68 & BUAA~\cite{BUAA20ReID} & 0.6522 \\
\hline
N/A (\textit{General}) & 75 & UAlbany~\cite{UAlbany20CountReIDMTMCTAnomaly} & 0.0368 \\
\hline
\end{tabular}
\end{table}

Most leading approaches utilized the provided synthetic data to enhance ReID performance through domain adaptation. Some of the methods trained real data with synthetic data by applying style transformation and content manipulation~\cite{Baidu20ReID, ZJU20ReID}. Other methods~\cite{RuiyanAI20ReID, Fraunhofer20ReID, NTU20ReID, Modulabs20ReID, UAlbany20CountReIDMTMCTAnomaly}, instead, trained classifiers for vehicle type, color, and viewpoint/orientation using the labels on synthetic data and made predictions on real-world data. Some teams~\cite{RuiyanAI20ReID, CUMT20ReID} also made use of identity clustering to generate pseudo-labels on the test data to expand the training set. Inspired by the state-of-the-art in person ReID, the methods with top performance in Track 2~\cite{Baidu20ReID, RuiyanAI20ReID, ZJU20ReID} all used ResNet with IBN structure as the backbone and applied pooling schemes including GMP, GAP, AAP, and AMP. Most teams combined identity classification (cross-entropy loss) and metric learning (triplet loss, circle loss, center loss, \textit{etc}.) in their training setup, \textit{e.g.},~\cite{Baidu20ReID, ZJU20ReID, NTU20ReID, BUPT20ReID, TUE20ReID, HCMUS20CountReIDAnomaly, Modulabs20ReID, CUMT20ReID, BUAA20ReID}. We have also seen various spatial, temporal and channel-wise attention mechanisms being utilized in methods such as~\cite{Fraunhofer20ReID, UMD20ReIDMTMCTAnomaly, NTU20ReID, BUPT20ReID, UAlbany20CountReIDMTMCTAnomaly}. Finally, it was shown in many algorithms~\cite{Baidu20ReID, RuiyanAI20ReID, ZJU20ReID, Fraunhofer20ReID, CUMT20ReID, BUAA20ReID} that re-ranking and other post-processing strategies were effective in improving the robustness of ReID. 

\subsection{Summary for the Track 3 challenge}

\begin{table}[t]
\caption{Summary of the Track 3 leader board.}
\label{table:3}
\centering
\footnotesize
\begin{tabular}{|c|c|c|c|}
\hline
Rank & Team ID & Team name (and paper) & Score \\
\hline\hline
1 & 92 & CMU~\cite{CMU20MTMCT} & 0.4585 \\
\hline
2 & 11 & XJTU~\cite{XJTU20MTMCT} & 0.4400 \\
\hline
5 & 72 & UMD~\cite{UMD20ReIDMTMCTAnomaly} & 0.1245 \\
\hline
6 & 75 & UAlbany~\cite{UAlbany20CountReIDMTMCTAnomaly} & 0.0620 \\
\hline
\end{tabular}
\vspace{-0.4cm}
\end{table}

All the teams followed the processing pipeline of object detection, multi-target single-camera (MTSC) tracking, ReID for appearance feature extraction, and spatio-temporal association to assign identities to tracklets across multiple cameras. The two best performing teams, \textit{i.e.}, CMU~\cite{CMU20MTMCT} and XJTU~\cite{XJTU20MTMCT}, both exploited metric learning and identity classification to train their feature extractors. Instead of explicitly associating targets in spatio-temporal domain, the team from XJTU~\cite{XJTU20MTMCT} embedded the information in an attention module and performed graph clustering based on pre-defined traffic topology. Similarly, the team from UMD~\cite{UMD20ReIDMTMCTAnomaly} built distance matrix using appearance and temporal cues to cluster tracks in multiple cameras. The UAlbany team~\cite{UAlbany20CountReIDMTMCTAnomaly} proposed a multi-camera tracking network that jointly learned appearance and physical features. 

\subsection{Summary for the Track 4 challenge}

\begin{table}[t]
\caption{Summary of the Track 4 leader board.}
\label{table:4}
\centering
\footnotesize
\begin{tabular}{|c|c|c|c|}
\hline
Rank & Team ID & Team name (and paper) & Score \\
\hline\hline
1 & 113 & Baidu-SYSU~\cite{Baidu20Anomaly} & 0.9695 \\
\hline
2 & 51 & USF~\cite{USF20Anomaly} & 0.5763 \\
\hline
3 & 106 & CET~\cite{CET20Anomaly} & 0.5438 \\
\hline
4 & 72 & UMD~\cite{UMD20ReIDMTMCTAnomaly} & 0.2952 \\
\hline\hline
N/A (\textit{General}) & 75 & UAlbany~\cite{UAlbany20CountReIDMTMCTAnomaly} & 0.9494 \\
\hline
N/A (\textit{General}) & 80 & HCMUS~\cite{HCMUS20CountReIDAnomaly} & 0.9059 \\
\hline
\end{tabular}
\vspace{-0.4cm}
\end{table}

The best performing Track 4 teams (\textit{i.e.,} Baidu-SYSU~\cite{Baidu20Anomaly} and USF~\cite{USF20Anomaly}) used a similar procedure: first pre-process and detect vehicles, then identify the anomalies, and finally perform a backtracking optimization to refine the anomaly prediction. The Baidu-SYSU team achieved an impressive prediction score of 0.9695. In their approach, they proposed a multi-granularity strategy, consisting of a box-level and a pixel-level tracking branch. The latter is inspired by the winning solution in AI City Challenge 2019~\cite{Bai19Traffic}. A fusion of the two strategies offers complementary views in anomaly refinement. The runner up proposed a fast, unsupervised system, where the anomaly prediction module used \textit{K}-means clustering to identify potential anomalous regions. The solution of the third-place team (CET~\cite{CET20Anomaly}) was also based on two complementary predictors: one works on the normal scale of videos, while the other works on a magnified scale on videos missed by the first predictor.

\section{Discussion}
\label{sec:discussion}

The 4th edition of the AI City Challenge has shown growing impact to the research communities, as the number of participants stayed strong and the quality of submissions was also highly improved. We summarize here several observations from the challenge results this year. 

The accuracy of vehicle counting depends highly on the quality of vehicle trajectory data. 
Challenges in this regard include the variety in camera views, image quality, lighting, and weather conditions. Participating teams have adopted state-of-art objection detection and tracking models to obtain vehicle trajectories. Among them, YOLOv3~\cite{Redmon18YOLOv3} and DeepSORT~\cite{Wojke17} were most widely used, and Faster R-CNN and Mask R-CNN~\cite{Ren15} were also popular.
Since most vehicles are traveling along the fixed traffic lanes, their motion pattern is predictable, and thus simple trackers based on IoU or linear motion-based trackers are effective. 
In the crowded and occluded scenarios, 
broken trajectories and identity switches can directly impact the counting accuracy. 
To this end, various post-processing methods were adopted.
To determine movement-specific vehicle counting, teams used both ROI-based and data-driven based MOI classification.
Both approaches require some level of camera-specific manual effort, and fully automatic methods are potential research topics in the future.
The winning team~\cite{Baidu20Count} has achieved over 0.95 counting accuracy. 
Lastly, for improving computational speed,
the team~\cite{CMU20Count} utilized frame-level parallelism and out-of-order-execution mechanisms for the bottle-neck detection stage with support for up to 8 GPUs. Many teams~\cite{CMU20Count, DiDi20Count, Baidu20ReID, VT20Count} have reported better than real-time processing speed.

Track 2 (vehicle ReID) is challenging due to two factors. First, vehicles present high intra-class variability caused by the dependency of shape and appearance on viewpoint. Second, vehicles also show small inter-class variability caused by the similar shape and appearance among vehicles produced by different manufacturers. The top performing teams in this task~\cite{Baidu20ReID, RuiyanAI20ReID, ZJU20ReID} built their algorithms based on state-of-the-art person ReID frameworks. Many models were trained on both identity classification loss and metric-learning-based loss that encouraged the network to maximally distinguish on fine-grained appearance features. Various attention mechanisms were also integrated to their proposed architectures to help the networks focus on representative information. Additionally, as we introduced augmented synthetic data in the challenge this year, many teams proposed to expand the training set with style-transformed simulated data, and learned models for classifying vehicle attributes and viewpoints using the automatically generated labels on these data. Another way teams used to gain additional data was to assign pseudo-labels to the test set based on clustering approaches. We anticipate that these types of methods will be used widely for real deployed systems, as manual annotation is costly and time-consuming. 

Vehicle ReID can be considered as a sub-task for Track 3 on city-scale MTMC vehicle tracking, where the algorithms not only need to learn discriminative appearance features for different identities, but also make use of spatio-temporal cues to associate targets across cameras at multiple intersections. The team from XJTU~\cite{XJTU20MTMCT} proposed a spatio-temporal attention module that learned the traveling time across adjacent cameras. They also introduced graph clustering in a distance matrix for grouping vehicle instances into continuous trajectories. The UMD team~\cite{UMD20ReIDMTMCTAnomaly} also utilized a similar approach for clustering tracks. In addition, teams applied state-of-the-art object detectors and MTSC tracking methods~\cite{Tang19MOANA, Tang18AIC18ICT, Tang17AIC17MultiKernelTrack, Lee18OnlineLearnICT} to generate reliable tracklets from each single camera. For instance, the top performing team from CMU~\cite{CMU20MTMCT} used Mask R-CNN~\cite{Ren15} and DeepSORT~\cite{Wojke17} for object detection and tracking, respectively. Compared to the ReID problem, MTMC tracking has more room for improvement before deployment in real world, especially as methods may not easily scale as the camera network grows. 

Traffic anomaly detection in Track 4 is challenging due to environmental factors, the complexity of the anomaly pattern, and insufficient anomaly training data. Since the use of external datasets were not allowed, teams thus mostly resorted to the provided training data for detector fine-tuning. The winning team (Baidu-SYSU) achieved very impressive prediction scores \cite{Baidu20Anomaly}. Their success is due to several notable reasons: (1) Instead of relying on a single-stage detector, they used two-stage Faster R-CNN \cite{ren2015faster} model with SENet \cite{hu2018squeeze} as the backbone. (2) They leveraged the experience from last year's winning model based on a pixel-level tracking branch in concert with a proposed box-level branch. This strategy can effectively improve the prediction accuracy. The runner up's solution was also interesting, offering competitive effectiveness with increased efficiency. The 4th AI City Challenge has successfully drawn the community's attention to this intriguing but challenging problem, and more effective solutions are yet to be explored in future research.

\section{Conclusion}

Through the AI City Challenge platform, we solicited original contributions in ITS and related areas where computer vision, and specifically deep learning, show promise in achieving large-scale practical deployment that will help make cities smarter. To accelerate the research and development of techniques, the 4th edition of this challenge presented three contributions: (1) The challenge introduced a track that not only measured effectiveness but also the computation efficiency. (2) Augmented synthetic data were introduced to substantially increase the number of training set samples for the ReID task, which could also be utilized in other tracks (3) Two leader boards are introduced in the evaluation system, where the \textit{Public} leader board was obtained from submissions without the use of external data, which encouraged contests close to the real-world use scenarios.
The 4th AI City Challenge has seen strong participation in all the four challenge tracks, where 76 out of 315 participating teams submitted their results and significantly improved the baselines on these challenging tasks. 

In the future, we will continue to push the state-of-the-art methods on real-world problems, by providing access to high-quality data and improving the evaluation platform. 


\section{Acknowledgement}

The datasets of the 4th AI City Challenge would not have been possible without significant contributions from an urban traffic agency in the United States and the DOT of Iowa. This challenge was also made possible by significant help from NVIDIA Corporation, which curated the datasets for public dissemination. 



{\small
\bibliographystyle{ieee_fullname}
\bibliography{aicity20,aicity19,aicity18,aicity17}
}

\end{document}